\documentclass[runningheads]{llncs}

\usepackage[T1]{fontenc}
\usepackage{graphicx,verbatim}
\usepackage{amsmath,amssymb,amsfonts}
\usepackage{xcolor}
\usepackage{booktabs}
\usepackage{tabularx}
\usepackage{nicefrac}
\usepackage{soul}
\usepackage[table,xcdraw, x11names]{xcolor}
\usepackage{multirow}
\usepackage{pifont}
\usepackage{xspace}
\usepackage{enumitem}
\usepackage{hyperref}

\newcommand\blfootnote[1]{%
  \begingroup
  \renewcommand\thefootnote{}%
  \footnotetext{#1}%
  \endgroup
}

\newcolumntype{Y}{>{\centering\arraybackslash}X}
\newcolumntype{Z}{>{\centering\arraybackslash}p{0.15\textwidth}} 

\newcommand*{\modelname}{BrainInterNet\xspace}
\newcommand{\cmark}{\ding{51}}
\newcommand{\xmark}{\ding{55}}

\begin{document}


\title{Interpretable Cross-Network Attention for Resting-State fMRI Representation Learning}

\titlerunning{Interpretable Cross-Network Attention for rs-fMRI}


\author{Karanpartap Singh\inst{1} \and
Adam Turnbull\inst{1} \and
Mohammad Abbasi\inst{1} \and
Kilian Pohl\inst{1} \and
Feng Vankee Lin\inst{1} \and
Ehsan Adeli\inst{1}
}
\authorrunning{Singh et al.}
%
\institute{Stanford University, Stanford, CA 94305 \\
\email{karanps@stanford.edu}}

  
\maketitle              


\begin{abstract}

Understanding how large-scale functional brain networks reorganize during cognitive decline remains a central challenge in neuroimaging. While recent self-supervised models have shown promise for learning representations from resting-state fMRI, their internal mechanisms are difficult to interpret, limiting mechanistic insight. We propose \textbf{\modelname}, a network-aware self-supervised framework based on masked reconstruction with cross-attention that explicitly models \textbf{inter-network} dependencies in rs-fMRI. By selectively masking predefined functional networks and reconstructing them from remaining context, our approach enables direct quantification of network predictability and interpretable analysis of cross-network interactions. We train \modelname on multi-cohort fMRI data (from the ABCD, HCP Development, HCP Young Adults, and HCP Aging datasets) and evaluate on the Alzheimer’s Disease Neuroimaging Initiative (ADNI) dataset, in total comprising 5,582 recordings. Our method reveals systematic alterations in the brain’s network interactions under AD, including in the default mode, limbic, and attention networks. In parallel, the learned representations support accurate Alzheimer’s-spectrum classification and yield a compact summary marker that tracks disease severity longitudinally. Together, these results demonstrate that network-guided masked modeling with cross-attention provides an interpretable and effective framework for characterizing functional reorganization in neurodegeneration.\blfootnote{Code published at \url{https://github.com/su-karanps/BrainInterNet}.}



\end{abstract}


\section{Introduction}
\label{sec:intro}

Resting-state functional magnetic resonance imaging (rs-fMRI) provides a powerful, noninvasive means of studying large-scale brain organization by capturing spontaneous, correlated fluctuations in neural activity \cite{Biswal1995FunctionalMri}. 
These intrinsic activity patterns reveal coherent functional networks that are remarkably consistent across individuals and have become a central object of study in systems and cognitive neuroscience \cite{Damoiseaux2006ConsistentSubjects,ThomasYeo2011TheConnectivity}. 
Accordingly, network-level interactions and functional connectivity have become central to understanding how the brain reorganizes across aging and neurodegenerative disease \cite{Bullmore2009ComplexSystems}.
Disruptions to this large-scale network organization are a hallmark of neurodegenerative disease, often preceding structural degeneration and clinical symptoms. 
In Alzheimer’s disease in particular, prior work has demonstrated selective vulnerability of functional systems, notably the default mode network, along with broader alterations in inter-network interactions as cognitive decline progresses \cite{Buckner2008iTheNetwork,Seeley2009NeurodegenerativeNetworks}. 

\begin{figure*}[t]
\centering
\includegraphics[width=1.0\textwidth]{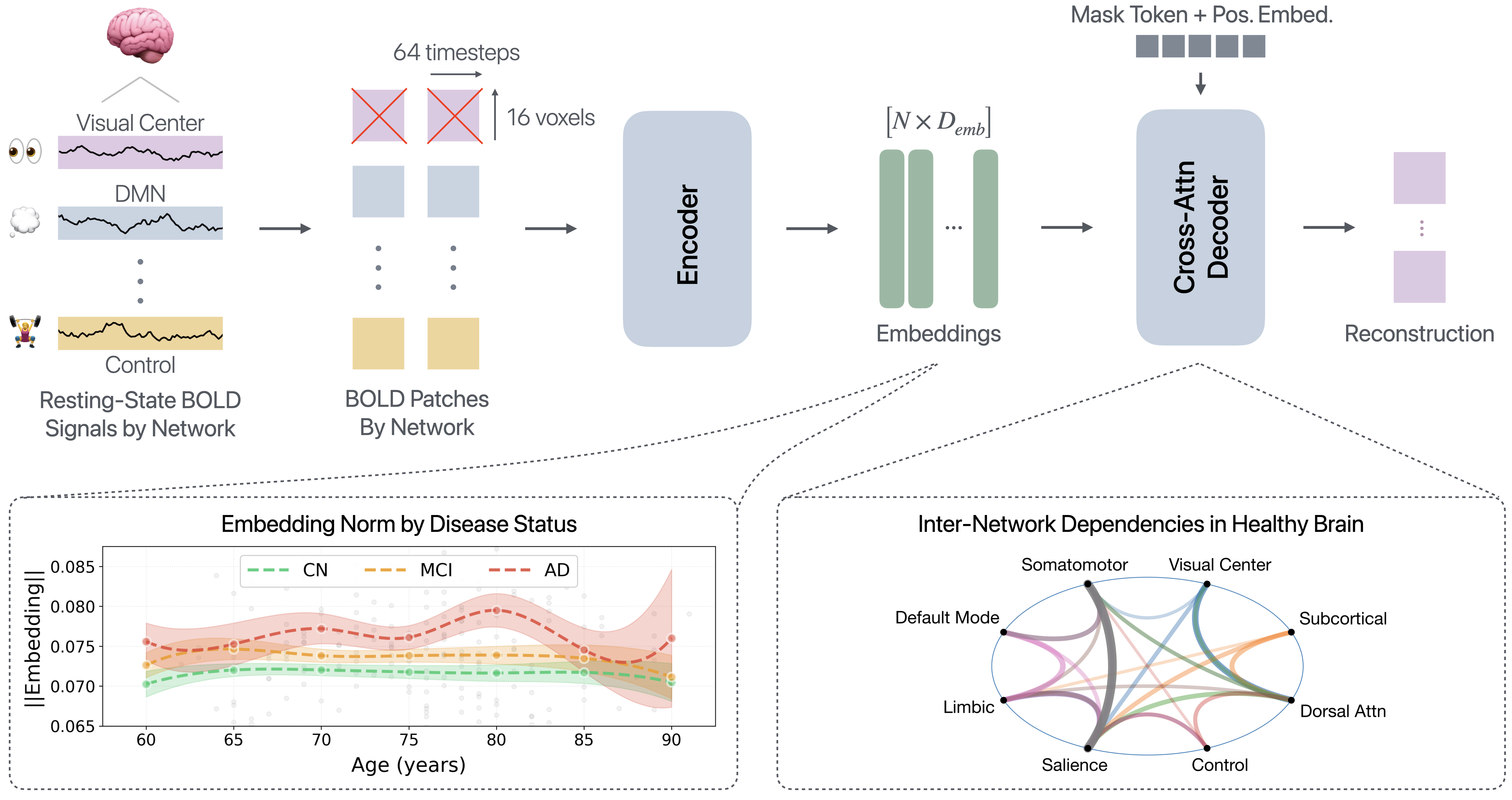}
\caption{\textbf{Overview of the proposed framework for modeling inter-network interactions in resting-state fMRI.} Resting-state BOLD signals are parcellated using the DiFuMo atlas and divided into non-overlapping temporal-spatial patches. Patches are grouped by Yeo functional networks \cite{ThomasYeo2011TheConnectivity}, and those belonging to a selected target network are masked. The encoder processes unmasked tokens to produce contextualized embeddings, while a cross-attention decoder reconstructs the masked network using the context and position embeddings. Learned embeddings reveal separation by disease status, offering a simple metric for mapping neurodegeneration trajectories (bottom left). Meanwhile, the cross-attention decoder provides interpretability into how predictions are made, offering a proxy for inter-network brain dependencies (bottom right).}
\label{fig:intro}
\end{figure*}

In parallel, recent advances in self-supervised learning have led to increasingly powerful models that leverage large unlabeled datasets to analyze rs-fMRI data, with an emphasis on predictive accuracy for tasks such as disease and demographic identification \cite{Caro2023BrainLM:Recordings,Dong2024Brain-JEPA:Masking,Yang2024BrainMass:Learning}. 
However, such models rely on dense latent embeddings whose internal structure is difficult to relate to known functional systems, limiting their utility for mechanistic analysis of brain network reorganization.
In this paper, we aim to bridge these perspectives by introducing a network-aware self-supervised framework that jointly supports accurate disease classification and interpretable analysis of inter-network dependencies. 
Our approach leverages functional network priors and a cross-attention architecture that work synergistically to model how functional networks predict one another in rs-fMRI, offering a model-intrinsic measure of inter-network brain organization that other approaches lack. 
Our contributions can be summarized as follows:

\begin{enumerate}[leftmargin=*, itemsep=8pt]
\item We introduce \modelname (Fig.~\ref{fig:intro}), a network-aware self-supervised framework that captures \emph{inter-network dependencies} in rs-fMRI by combining network-aligned tokenization and masking, and a cross-attention–only decoder.

\item We propose a principled reconstruction-based analysis of inter-network organization, where cross-attention–derived contribution profiles are intrinsic to the training objective, unlike post-hoc correlation or attribution analyses, and directly quantify how functional systems depend on one another.

\item We demonstrate that \modelname captures Alzheimer's disease-related functional reorganization, achieving competitive classification performance while revealing systematic network alterations that characterize disease progression.
\end{enumerate}


\section{Related Work}
\label{sec:literature}

Resting-state fMRI has been widely used to characterize large-scale functional brain organization and its alteration in aging and neurodegenerative disease.
Early work established that spontaneous low-frequency BOLD fluctuations form reproducible intrinsic connectivity networks, while subsequent connectivity and graph-theoretic studies showed that aging and Alzheimer’s disease are associated with reduced network segregation, disrupted default mode connectivity, and altered global topology \cite{Biswal1995FunctionalMri,Buckner2008iTheNetwork,Damoiseaux2012FunctionalDisease,Greicius2004Default-modeMRI,Stam2006Small-WorldDisease}.
These network-level alterations have been shown to correlate with cognitive impairment and disease severity, motivating their use as imaging biomarkers for studying disease progression.

In parallel, recent work has applied self-supervised learning at scale to rs-fMRI, producing versatile “foundation models” for brain connectivity \cite{Zhou2025BrainDiscovery}. These models demonstrate the effectiveness of large-scale predictive pretraining, achieving strong performance on disease classification, age and sex prediction, and cross-dataset transfer \cite{Caro2023BrainLM:Recordings,Dong2024Brain-JEPA:Masking,Yang2024BrainMass:Learning}. However, their primary emphasis remains representation learning and predictive accuracy, with limited emphasis on interpretability.
Although some studies report associations between learned features and canonical networks \cite{Caro2023BrainLM:Recordings}, these analyses are post-hoc and do not provide a principled account of how model predictions are made.
In contrast, we introduce a framework in which functional networks explicitly define the masking, reconstruction, and attention structure during training, yielding network-level predictability and contribution measures that are intrinsic to the model.
This enables mechanistic analysis of inter-network dependencies underlying model predictions, addressing a key limitation of existing foundation-model approaches.


\section{\modelname}
\label{sec:approach}

Our model builds on the CrossMAE architecture \cite{Fu2024RethinkingAutoencoders}, consisting of a transformer encoder and a cross-attention–only decoder trained in a self-supervised reconstruction setting (Figure~\ref{fig:intro}).
Raw rs-fMRI data are four-dimensional volumes (time $\times x \times y \times z$) with high dimensionality and noise, making voxel-level modeling computationally intractable and prone to overfitting in moderate-sized cohorts.
Following prior work, we therefore adopt a region-based representation using the DiFuMo atlas \cite{Dadi2020Fine-grainAnalysis}, which parcellates the cortex into 1024 functionally defined regions of interest (ROIs), substantially reducing dimensionality while preserving functional organization.
For each subject, we extract parcel-wise time series segments of shape $(T, C)$ with $T=64$ time points and $C=1024$ parcels, and divide them into non-overlapping temporal–spatial patches of size $(16,16)$.
This produces a sequence of tokens, each representing 16 consecutive time points from 16 ROIs, with ROIs ordered such that each token corresponds to a single functional network.
Tokens are linearly projected into a $D_{\text{emb}}$-dimensional embedding space and processed by the encoder.
During pretraining, all tokens corresponding to a randomly selected functional network are masked.
The encoder processes only unmasked tokens, while the cross-attention decoder reconstructs the masked network from the encoded context.

\subsection{Tokenization and Network-Guided Masking}



We organize the token sequence such that each functional network corresponds to a contiguous block of tokens.
Specifically, parcels are reordered according to their Yeo-network membership \cite{ThomasYeo2011TheConnectivity}, and minimal zero-padding is applied when necessary to ensure consistent alignment.
As a result, all tokens associated with a given network appear consecutively in the input sequence.
For example, if a network spans 48 parcels and each token represents 16 parcels, the network is represented by three consecutive tokens.
Additionally, rather than masking random tokens, we perform structured masking at the network level.
For each training sample, we randomly select one target network and mask all tokens associated with that network.
Let $\mathcal{N}=\{1,\dots,N\}$ denote the set of networks.
For a given masked network $i \in \mathcal{N}$, all tokens belonging to $i$ are removed from the encoder input.
The encoder therefore receives tokens from $\mathcal{N}\setminus\{i\}$, and the decoder is tasked with reconstructing the missing network from the remaining context. 
This network-guided masking strategy enables systematic analysis of how well each functional network can be predicted from others.
By construction, each reconstruction task corresponds to predicting the activity of one network conditioned on the rest, providing a direct probe of inter-network dependencies.

\subsection{Model Architecture and Training}

The encoder consists of $L_E$ transformer blocks with multi-head self-attention and feedforward layers.
It processes only unmasked tokens and produces embeddings $\mathbf{Z} \in \mathbb{R}^{N_u \times D_{\text{emb}}}$, where $N_u$ is the number of unmasked tokens. 
The decoder receives a learnable mask token and positional embeddings corresponding to the masked network.
It applies $L_D$ layers of cross-attention, where masked queries attend only to encoder outputs $\mathbf{Z}$ with no self-attention.
This design enforces that reconstruction of masked networks relies solely on information encoded from the remaining networks, yielding an explicit dependency structure.
The model is trained to minimize an MSE loss between predicted and ground-truth patches.



\section{Experiments}
\label{sec:experiments}

We use $N_\text{tr}=3,087$ resting-state fMRI BOLD recordings for pre-training \modelname from the Human Connectome Project Young-Adults (HCP-YA) \cite{VanEssen2013TheOverview}, HCP-Aging \cite{Bookheimer2019TheOverview}, HCP-Development \cite{Somerville2018TheOlds}, and ABCD \cite{Casey2018TheSites} datasets. We then use $N_\text{ts}=2,366$ recordings from ADNI \cite{Petersen2010AlzheimersADNI} as a downstream dataset to demonstrate the capability of our approach in modeling neurodegeneration. 
Each dataset provides whole-brain BOLD signals acquired under resting conditions, from which we extract parcel-wise time series for each subject.
We reserve 80\% of the data for training and 10\% each for validation and testing, sample ten random 64-timestep segments per recording, and train for 50 epochs with a warmup-stable-decay learning-rate schedule \cite{Wen2024UnderstandingPerspective}. The datasets are partitioned at the subject level to prevent any repeated or longitudinal recordings from the same individual from appearing in both training and testing sets. 
All datasets are preprocessed using standardized pipelines \cite{Bookheimer2019TheOverview,Somerville2018TheOlds,Abbasi2025FMRIControl,Esteban2019FMRIPrep:MRI}.

Our framework supports two complementary analyses: (i) global characterization of learned representations via encoder embeddings, and (ii) network-level analysis of reconstruction dynamics via the cross-attention decoder. The encoder embeddings provide a compact summary of subject- and group-level variation that captures disease-related trajectories, while the decoder enables explicit analysis of inter-network predictability and network-specific dependencies. We describe these analyses in turn in the following sections.

\begin{figure*}[t]
\centering
\includegraphics[width=1.0\textwidth]{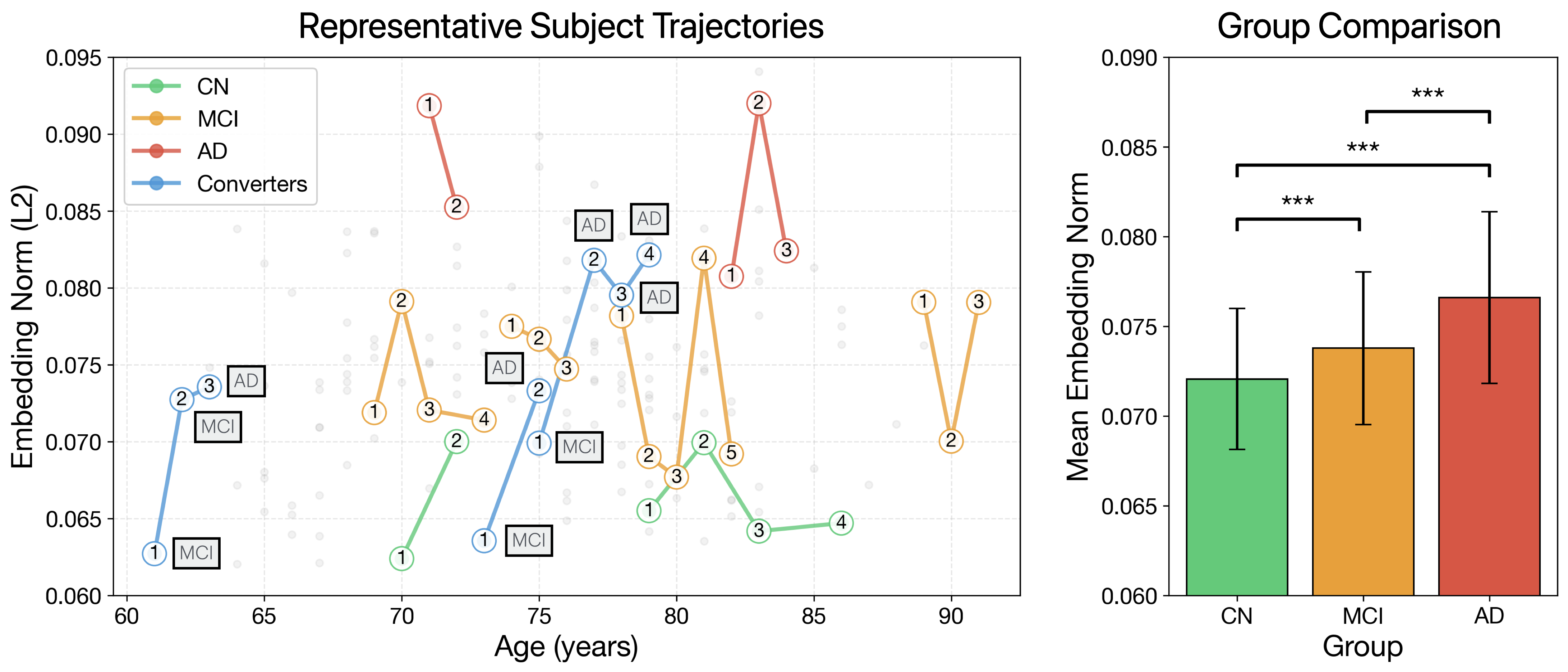}
\caption{\textbf{Longitudinal charting of Alzheimer’s disease progression using encoder embeddings.} 
\textbf{Left:} Representative longitudinal trajectories of the encoder embedding norm ($\ell_2$) for individual subjects, including stable CN, stable MCI, stable AD, and converters who transition from MCI to AD across sessions. Faint background points show embedding norms for all subjects and sessions. 
\textbf{Right:} Group-level comparison of embedding norms for stable CN, MCI, and AD subjects, showing progressive separation with disease severity (mean $\pm$ s.e.m.). *** indicates $p<0.001$ (two-sided Welch's t-test).
Together, these analyses illustrate that the learned encoder representations capture both group-level differences and subject-specific trajectories associated with cognitive decline.
}
\label{fig:longitudinal}
\end{figure*}

\subsection{Encoder Embedding Trajectories and Disease Progression}

Beyond their use for downstream tasks, the learned encoder representations exhibit structured variation with age and clinical status.
As shown in Fig.~\ref{fig:intro}, group-averaged embedding norms reveal clear separation between cognitively normal (CN), mild cognitive impairment (MCI), and Alzheimer’s disease (AD) cohorts.
While CN and MCI groups exhibit relatively stable trajectories with age, AD subjects show a pronounced elevation in embedding magnitude, with a peak in the late 70s followed by a decline at older ages, remaining consistently above CN and MCI. To further assess whether these group-level trends reflect subject-specific disease trajectories, we examine longitudinal sessions for individual participants (Fig.~\ref{fig:longitudinal}).
CN and MCI subjects exhibit stable embedding trajectories across sessions and ages, whereas AD subjects maintain elevated embedding norms over time.
Notably, subjects who convert from CN or MCI to AD display marked increases in embedding magnitude coincident with clinical progression, indicating sensitivity to disease progression.
Consistent with these observations, group-level comparisons of stable subjects show significant differences in embedding norm between CN, MCI, and AD cohorts (Fig.~\ref{fig:longitudinal}), indicating that this global representation statistic tracks disease severity.

\begin{figure*}[t]
\centering
\includegraphics[width=1.0\textwidth]{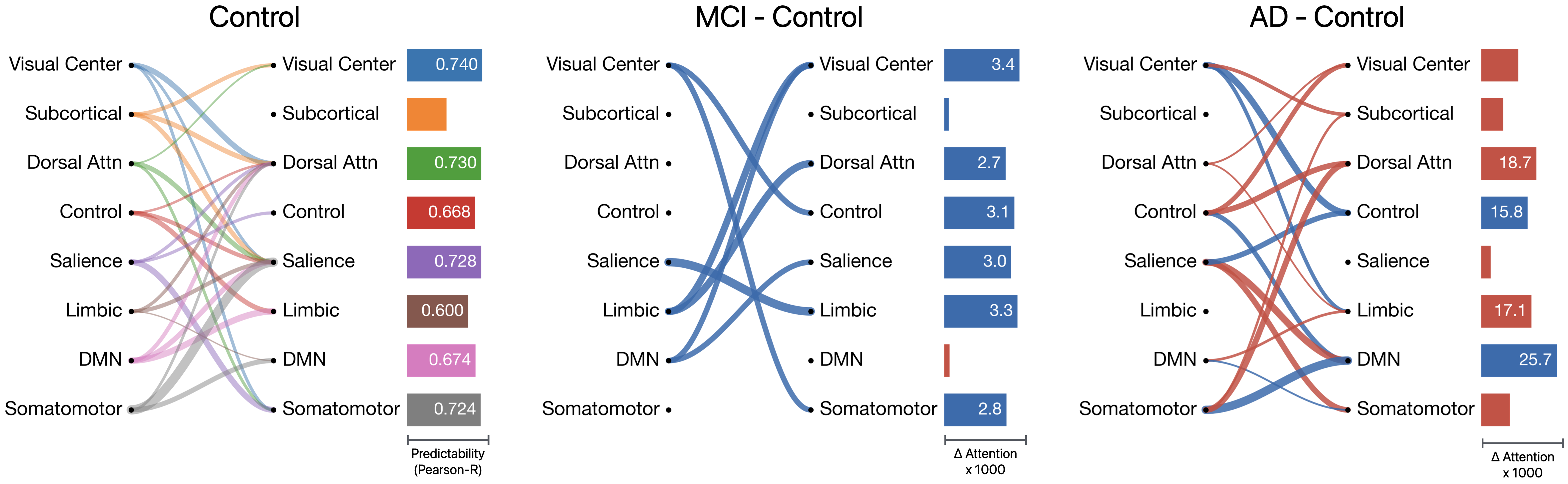}
\caption{\textbf{Inter-network predictability and disease-related changes in functional coupling}.
\textbf{Control:} Baseline inter-network predictability in cognitively normal (CN) subjects. Chord diagrams depict cross-network attention-based contributions during reconstruction, where edge thickness indicates the \textit{relative} influence of source networks (left) on target network prediction (right). Bar plots report network-wise predictability measured by Pearson correlation between reconstructed and true signals.
\textbf{MCI - Control:} Changes in inter-network contributions from CN to MCI. Edges and bars represent \textit{differences} in attention-derived source dependencies relative to CN.
\textbf{AD - Control:} Analogous changes in inter-network contributions from CN to AD. \textcolor{red!70!black}{Red} indicates increased reliance on a given source network for reconstruction, while \textcolor{blue!70!black}{blue} indicates decreased reliance.
Together, these panels illustrate progressive reorganization of functional dependencies from cognitive impairment and Alzheimer’s disease.}
\label{fig:attention}
\end{figure*}

\subsection{Inter-Network Dependence \& Neurodegeneration}

To quantify inter-network dependencies, we evaluate the predictability of each functional network by masking all tokens associated with a target network and measuring reconstruction accuracy from the remaining context. Predictability is quantified using the Pearson correlation between predicted and true signals, averaged across tokens and subjects from held-out data, yielding a network-wise profile of how well each system can be inferred from others. To characterize how these reconstructions are formed, we analyze decoder cross-attention weights. For each masked network, cross-attention weights are aggregated across the top 12 high-magnitude heads, which together account for the majority of cross-attention mass, while remaining heads contribute marginal and diffuse weights. These weights are then grouped by Yeo network, producing a contribution vector that reflects the relative influence of each unmasked network during reconstruction.

Figure~\ref{fig:attention} summarizes baseline inter-network predictability in cognitively normal (CN) subjects and disease-related changes across the Alzheimer’s disease spectrum. In CN individuals, network reconstruction relies on a distributed pattern of cross-network contributions, with relatively balanced predictability across sensory, attentional, and association systems. Relative to CN, MCI is associated with selective shifts in contributions and modest changes in predictability, consistent with stage-specific reweighting rather than global degradation. In Alzheimer’s, these alterations become more pronounced, with larger redistribution and corresponding changes in predictability, particularly involving the default mode, limbic, and dorsal attention networks.

\begin{table*}[h]
\centering
\caption{Performance comparison on 3-way Alzheimer’s disease spectrum classification on ADNI. Reported metrics including balanced accuracy (BAcc), F1-score, and AUC-ROC. We test four baselines, and three ablations: a vanilla MAE, replacing cross-attention with self-attention, as well as replacing the network-based masking (denoted Net in the table) with random masking (denoted Rand). All models are finetuned.}
\label{tab:adni_results}
\begin{tabularx}{\textwidth}{l l Z Z Y Y Y}
\toprule
& \textbf{Method} & \textbf{Mask} & \textbf{CrossAttn} & \textbf{BAcc (\%)} & \textbf{F1 (\%)} & \textbf{AUC} \\ \midrule

\multirow{4}{*}{\rotatebox{90}{\textbf{\tiny Baselines}}}
& BrainLM  \cite{Caro2023BrainLM:Recordings}   & Rand & \xmark & 68.78 & 68.70 & 0.776 \\
& Brain-JEPA \cite{Dong2024Brain-JEPA:Masking} & Rand & \xmark & 51.85 & 51.10 & 0.620 \\
& BrainGNN  \cite{Li2021BrainGNN:Analysis}  & NA & \xmark & 40.89 & 40.22 & 0.553 \\ 
& Random & NA & \xmark & 33.65 & 31.50 & 0.503 \\
\midrule

\multirow{2}{*}{\rotatebox{90}{\textbf{\tiny Ablations}}}
& MAE \cite{He2021MaskedLearners} & Rand & \xmark & 69.31 & 71.86 & 0.849 \\
& Ours - Net & Rand & \cmark & 48.26 & 43.23 & 0.520 \\
& Ours - Cross         & Net & \xmark & 60.29 & 58.22 & 0.615 \\ \midrule

& \textbf{\modelname} & Net & \cmark & \textbf{77.53} & \textbf{77.55} & \textbf{0.850} \\

\bottomrule
\end{tabularx}
\end{table*}

\subsection{Downstream Classification}

For Alzheimer’s-spectrum classification, we attach a lightweight prediction head to the pretrained encoder and fine-tune the model end-to-end on labeled ADNI data using \textbf{subject-level} data splits. Encoder outputs are pooled and passed to a shallow multi-layer perceptron, and all baselines are reproduced under identical splits, optimization schedules, and early stopping criteria. As shown in Table~\ref{tab:adni_results}, our network-aware cross-attention model achieves strong balanced accuracy, F1-score, and AUC-ROC on three-way CN/MCI/AD classification, outperforming or matching recent foundation-model baselines \cite{Caro2023BrainLM:Recordings,Dong2024Brain-JEPA:Masking} and task-specific models \cite{Li2021BrainGNN:Analysis}, despite being trained on substantially smaller datasets. Notably, this three-way setting is considerably more challenging than the binary classification tasks commonly reported in prior work due to the heterogeneity and intermediate nature of MCI. In this regime, representations optimized for global separability may miss differences characteristic of neurodegeneration, whereas explicitly modeling inter-network dependencies captures graded functional reorganization.


\section{Discussion}
\label{sec:discussion}

We presented \modelname, a network-aware self-supervised framework for modeling resting-state fMRI that integrates structured network-level masking with cross-attention-based reconstruction to enable interpretable analysis of inter-network dependencies.
Applied to Alzheimer’s disease, \modelname captures both global representational changes and network-specific coupling alterations across disease stages.
Encoder embeddings provide a compact summary marker that separates CN, MCI, and AD groups and reflects longitudinal progression at the subject level, while the cross-attention decoder reveals brain functional network dependencies.
These results align with established findings that disease-related changes do not reflect uniform signal degradation but instead involve selective reweighting of functional dependencies between networks \cite{Brier2012LossProgression}.

Specifically, we replicated the finding that AD predominantly involves changes to the DMN and other association networks (e.g., limbic and control) that are early to be affected by AD pathology \cite{Greicius2004Default-modeMRI,Seeley2009NeurodegenerativeNetworks}.
These changes reflect both increases and decreases in inter-network dependencies, replicating broad alterations in large-scale network topology in AD.
On the other hand, changes in MCI were comparatively subtle, in line with the known heterogeneity and less clear network re-organization in this population \cite{Damoiseaux2008ReducedAging,Gauthier2006MildImpairment}. 
By explicitly modeling how networks predict one another, our framework builds on traditional connectivity and graph-theory/topology analyses, leveraging the increased power of large-scale models to detect non-linear dependencies while maintaining interpretability.

Future work should aim to more comprehensively compare the findings of this approach with traditional measures, particularly considering the role of directionality and non-linearity. Additionally, although this work focuses on AD as a representative application, the proposed methodology is broadly applicable to other neurodegenerative and neuropsychiatric conditions characterized by network-level dysfunction.

\newpage



%
%
%
\bibliographystyle{splncs04}
\bibliography{references}

\end{document}